\pdfoutput=1
\documentclass{article}
\usepackage{preprint,times}
\usepackage{amsmath,amssymb,mathtools}
\usepackage{graphicx}
\usepackage{booktabs}
\usepackage{array}
\usepackage{multirow}
\usepackage{xcolor}
\usepackage{hyperref}
\usepackage{url}
\usepackage{microtype}
\usepackage{enumitem}
\usepackage{makecell}
\usepackage{float}

\title{Concepts Are Distributions: Training-Free\\Insertion in Vision-Language Models}
\author{Deblina Bhattacharjee\\
Department of Computer Science\\
University of Bath}

\newcommand{\inject}{\textsc{Injection}}
\newcommand{\remove}{\textsc{Removal}}
\newcommand{\retain}{\textsc{Retention}}
\newcommand{\fullrep}{\textsc{Full replacement}}

\hypersetup{hidelinks}

\begin{document}
\maketitle

\begin{abstract}
Many activation-steering methods represent a concept as a direction in activation space. We test whether this representation also supports object insertion in the visual tokens of frozen vision-language models (VLMs). We find a clear asymmetry. A single direction can suppress an object, but it does not reliably introduce an object that is absent from the image. Learned directions, text-derived directions, and individual real activations all perform poorly. We show that the limitation is distributional. Repeating one concept vector across visual tokens can match the mean while collapsing covariance. By contrast, sampling the tokens from a Gaussian whose mean and covariance match those of real object activations performs on par with transplanting a real object. We therefore store each concept as a low-rank Gaussian estimated from a few images. On CLIP-L/14, four covariance directions are sufficient, giving a 20~KB representation per concept. Other vision encoders require modestly higher rank. The method requires no training and no reference image at inference time, can be used without modification with three language models that share a vision encoder, extends to two further models with different encoders, and works without object-location annotations. Our results show that the covariance first makes the concept mean readable and that its concept-specific part then contributes identity. This suggests that, for object insertion, a concept in the visual stream is better described by a distribution over tokens than by a single direction.
\end{abstract}

\section{Introduction}

Vision-language models generate text from visual tokens produced by an image encoder~\citep{liu2023}. Since each token corresponds to an image patch, the tokens covering an object can be edited directly, without changing the input image or any model weight. Furthermore, existing steering methods for VLMs operate on the same tokens, and our findings also hold at three other points of the model (Appendix~\ref{app:robust}). In language models, many concepts can be approximated by directions in activation space, and adding or subtracting such directions can induce or suppress the corresponding behavior~\citep{arditi2024}. The same view has been adopted for VLMs. Visual-token directions have been used to reduce hallucination~\citep{li2025hidden} and to remove objects from generated captions~\citep{jiang2025}.

\begin{figure}[t]
    \centering
    \includegraphics[width=0.78\linewidth]{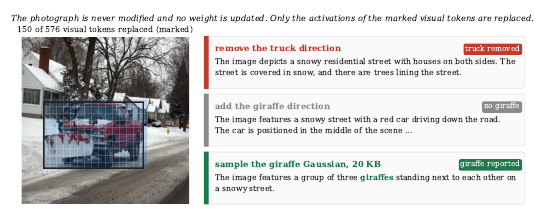}
    \caption{Overview on one image. The photograph and all model weights remain unchanged. We edit only the marked visual tokens before they reach the language model. Projecting out the truck direction removes the truck from the caption. Adding the giraffe direction to the same tokens does not make the model report a giraffe. Replacing those tokens with samples from the stored 20~KB Gaussian for the giraffe concept makes the model report a giraffe.}
    \label{fig:overview}
\end{figure}

We ask whether this directional view also supports the converse operation: introducing a concept that is absent from an image. We find that it does not. Figure~\ref{fig:overview} illustrates the asymmetry. Projecting an object direction out of the visual tokens that cover an object removes it from the caption in a substantial fraction of cases. By contrast, adding the direction of a target concept at the same positions is considerably less effective. In particular, the strongest single direction that we tested, namely, a tuned text-derived direction~\citep{jiang2025}, introduces the target in 32.0\% of the cases. Other single vectors do not perform better, including the real activation that best represents the concept and a direction sampled from a Gaussian over concept vectors~\citep{zhao2025}. These interventions can remove the original object, but they generally do not create the target.

Intuitively, removing an object requires disrupting its representation, which many perturbations achieve. By contrast, inserting an object requires producing tokens that the language model interprets as an object. Since the tokens of a real object vary from patch to patch in a structured manner, a single direction cannot provide them. An object occupies $m$ visual tokens. Replacing those tokens with $m$ copies of one vector produces a set whose mean is that vector and whose tokens all share the same direction. Adding one vector instead preserves only the covariance already present in the source tokens. We test the role of second-order structure with surrogates that preserve progressively more statistics of a real target object. Preserving only the mean yields 6.7\% injection. Preserving every per-dimension marginal while destroying covariance reaches 20.0\%. By contrast, a Gaussian whose population mean and covariance match the object reaches 52.7\%, even though it contains no real activation. This is statistically indistinguishable from transplanting the real object at 50.0\% (11:15, $p=0.56$). Here, $a$:$b$ denotes the number of images on which the first condition succeeds while the second fails, and conversely, with $p$ obtained from an exact McNemar test. Note that this bounds the difference at approximately 0.1 but does not establish equality. Existing feature-transfer operators, including AdaIN~\citep{huang2017adain} and WCT~\citep{li2017wct}, follow the same ordering when ranked by the target second-order structure they transfer.

This observation leads directly to a practical representation. We estimate the mean and covariance of a concept once from a small set of images and store a low-rank factorization. At inference time, we sample from these statistics instead of requiring a reference image or a stored real activation. On CLIP-L/14, rank four costs 20~KB per concept and reaches 59.3\% injection, exceeding a real donor object (28:14, $p=0.044$). The same stored statistics can be used with three language models that share the vision encoder, and we observe the same effect on two models with different encoders. Direct questioning gives the same qualitative conclusion. The unedited model answers ``no'' for the absent target in 98.0\% of cases, whereas the edited model answers ``yes'' in 88.0\%. Under the same edit, it continues to reject an unrelated absent concept.

Our results support training-free counterfactual interventions in multimodal representations and clarify why constructive editing differs from suppression (Section~\ref{sec:removal}). Our main contributions are:
\begin{itemize}[leftmargin=1.6em]
    \item We identify and explain an asymmetry between removing and inserting concepts in the visual-token stream of a frozen VLM. Removal succeeds with a rank-one modification, whereas insertion depends on second-order structure, in the sense that the covariance makes the concept mean readable and its concept-specific part contributes identity.
    \item We show that real activations are unnecessary for insertion. A low-rank Gaussian estimated from a few images matches or exceeds activation transplantation across five models and three vision encoders.
    \item We verify the edit with direct questions, free recall, forced choice, and human judgment, showing that the effect is specific to the inserted concept and is not a generic tendency to affirm.
\end{itemize}
Additional controls, null results, and failed predictions are reported in Appendix~\ref{app:controls}.

\section{Related Work}

\paragraph{Concepts as directions.}
Steering vectors, refusal directions, and task vectors represent concepts as directions in activation space~\citep{arditi2024,li2025hidden,huang2024task}. The closest work to ours is GCS~\citep{zhao2025}, which models uncertainty over probe-derived concept vectors with a Gaussian but still adds one sampled direction to the representation. Our Gaussian instead models the distribution of token activations, and an intervention substitutes $m$ samples from that distribution. A single draw from a concept-vector Gaussian behaves like the other single-vector operators in our experiments. \citet{birardi2025} also report that suppression is easier than targeted steering for attribution-graph features in language models, which they attribute to feature coverage. We identify a different mechanism in the vision-language setting.

\paragraph{Second-order statistics in feature transfer.}
Neural style transfer has represented style through Gram matrices since \citet{gatys2016}. AdaIN matches per-channel means and variances~\citep{huang2017adain}, while WCT transfers the full covariance~\citep{li2017wct}. We use these operators as controlled probes of object identity in a frozen VLM. This moves the question from image synthesis to language readout and from matching style statistics to synthesizing object-token distributions.

\paragraph{Editing VLM representations.}
ProjectAway removes objects without training by orthogonalizing image features against a text embedding~\citep{jiang2025}. VISTA-SAE~\citep{kravets2026}\footnote{We use ``VISTA-SAE'' to distinguish the SAE-transfer method of \citet{kravets2026} from the unrelated VISTA method of \citet{li2025hidden}.} performs localized removal and replacement after training a projector for roughly two hundred GPU-hours. Our method is training-free and supports both removal and insertion.

\paragraph{Class-conditional features, activation patching, and dictionaries.}
Class-conditional Gaussian models are standard in anomaly detection~\citep{lee2018,defard2021}. Our shared-covariance variant uses the same statistical model generatively. Activation patching is an established interpretability primitive~\citep{vig2020,meng2022,heimersheim2024}, and \citet{huang2024task} use mean activations constructively. Our analysis suggests that a tiled mean should be insufficient for insertion. Sparse autoencoders provide another route to visual features, but they require additional training~\citep{joseph2025,pach2025,leask2025,klindt2023}. Archetypal variants constrain learned atoms to the convex hull of the data~\citep{fel2025}, whereas Gaussian samples are not subject to this constraint.

\section{Method}
In this section, we first define the intervention (Section~\ref{sec:setting}). We then introduce the surrogates (Section~\ref{sec:surrogates}), which serve as diagnostic tools in Section~\ref{sec:requirements} to determine the statistics that insertion depends on, and finally present the resulting method (Section~\ref{sec:bankmethod}).

\subsection{Problem setting}
\label{sec:setting}
We consider a frozen VLM whose vision encoder produces patch activations $h_1,\ldots,h_N\in\mathbb{R}^d$. These activations pass through a projector before they are read by the language model. Let $P$ denote the set of patches covering an object bounding box and let $m=|P|$. An insertion intervention replaces $\{h_p\}_{p\in P}$ with $m$ substitute vectors. Each substitute $x_p$ is scaled to the norm of the activation it replaces, such that the token written at position $p$ is $s_p x_p$, with $s_p=\lVert h_p\rVert/\lVert x_p\rVert$. No other part of the forward pass changes and no weight is updated. We call the resulting $m\times d$ matrix the \emph{substituted set} and ask which statistics of this set determine whether the language model reports a target concept.

\subsection{Moment-matched surrogates}
\label{sec:surrogates}
To isolate the relevant statistics, we construct surrogates of a real donor block $D\in\mathbb{R}^{m\times d}$ taken from an image of the target class. Let $\mu$ be the row mean of $D$, let $d_i$ denote its rows, and define $X=D-\mu$.

The \emph{mean} surrogate tiles $\mu$ over all $m$ patches. Its first moment is exact and its covariance is zero before norm matching. After norm matching, the tokens $s_p\mu$ differ in norm but remain collinear, and the set thus has no variation outside the mean direction. The \emph{marginals} surrogate independently permutes every feature dimension across rows. It therefore preserves every per-dimension marginal and contains exactly the same $md$ numerical values as $D$, while destroying all dependence between dimensions, including their covariance.

For the \emph{Gaussian} surrogate, we draw
\begin{equation}
    x_j = \mu + \sum_i c_{ij}(d_i-\mu),
    \qquad c_{ij}\sim\mathcal{N}(0,1/m),
    \label{eq:gaussian-surrogate}
\end{equation}
and recenter the samples so that their empirical mean is $\mu$. The population covariance of the sampling distribution is $X^\top X/m$. A finite set of $m$ samples matches this covariance in expectation. Note that norm matching rescales each sample by a positive factor and thus cannot create a new direction. In Appendix~\ref{app:norm}, we show that the results remain unchanged under two other normalizations. No sample is a real activation. Finally, the \emph{four-object} surrogate draws real target-class activations from four images. This preserves class-level marginals without preserving the joint structure of a single object.

\begin{figure}[t]
    \centering
    \includegraphics[width=0.9\linewidth]{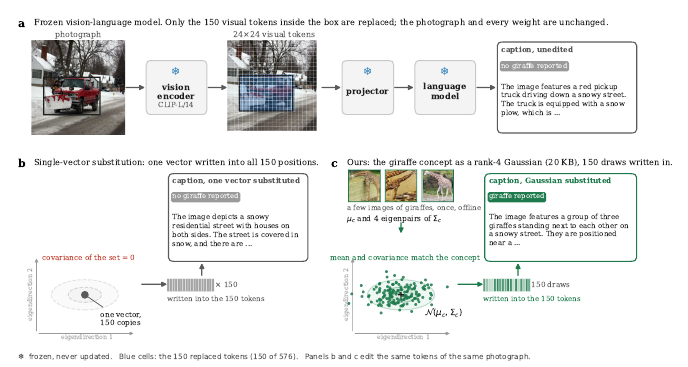}
    \caption{Intervention on one image. The frozen model encodes the photograph into a $24\times24$ grid of visual tokens. We replace only the tokens inside the marked box. Repeating one vector across these positions gives a substituted set with no variation in direction and does not make the model report the target. Our method estimates a low-rank Gaussian for the target concept from a few images and writes $m$ independent draws into the same positions. The model then reports the target.}
    \label{fig:method}
\end{figure}

\subsection{Gaussian concept bank}
\label{sec:bankmethod}
If a concept is adequately described by the mean and covariance of its token distribution, these statistics can be estimated offline and stored once. For each class $c$, we pool the activations inside the bounding boxes of a small number of images. We store the mean $\mu_c$ together with the leading $r$ eigenpairs of the covariance $\Sigma_c$. At edit time, we draw $m$ activations from the resulting rank-$r$ Gaussian and substitute them at the positions in $P$ (Figure~\ref{fig:method}). The storage cost is $(r+1)d$ values per concept. For CLIP-L/14 with $r=4$, this is 20~KB. The procedure requires no training, no segmentation model, and no clustered dictionary.

\section{Experiments and Results}

\subsection{Experimental setup}
Unless stated otherwise, we intervene on a frozen LLaVA-1.5-7B~\citep{liu2023} at the input of its multimodal projector. The visual stream contains $N=576$ patch activations of dimension $d=1024$ from a CLIP-L/14-336 vision tower~\citep{radford2021}. We use images from the COCO validation set~\citep{lin2014} in which the source object occupies 2--30\% of the frame. Removal can only be scored when the unedited caption names the source object. Of 400 eligible images, 286 satisfy this criterion. We use the first 150 for condition sweeps and reserve the remaining 136 as held-out recognized images. The 114 images on which the source is not named are later used to test insertion outside the selected population.

We report four quantities. \inject{} is the proportion of edited captions that name the target concept. We validate this metric against direct questioning in Section~\ref{sec:direct} and against human judgment in Appendix~\ref{app:human}. \remove{} is the proportion of edited captions in which the source object is no longer named. \retain{} is the proportion of content words from the unedited caption, excluding the source and target concepts, that remain in the edited caption. We use it as an automatic proxy for scene fidelity. \fullrep{} is the proportion of captions that name the target and no longer name the source. Every condition within an experiment uses the same images and target assignments, so comparisons are paired. Gaussian and donor conditions are resampled for each experiment. Their absolute rates therefore vary by up to 0.02 across experiments (seed s.d. 0.016), and each comparison reports the values of the run it belongs to. We use exact McNemar tests and report discordant counts as $a$:$b$, where $a$ is the number of images on which the first condition succeeds and the second fails, and $b$ is the converse. Appendix~\ref{app:protocol} gives the complete protocol.

\subsection{What insertion requires}
\label{sec:requirements}

In this section, we study which statistics of the substituted set insertion depends on. We first test whether the provenance of a single vector matters. Substituting one vector at every position in $P$ gives a set with mean $v$ and no variation in direction. Adding the same $v$ to each token instead preserves the covariance of the source tokens. Neither operation supplies the second-order structure of the target. Adding a target direction obtained from a clustered visual dictionary gives 13.8\% injection. Overwriting the same tokens with that direction gives 11.3\%. A medoid target activation gives 22.6\%, an arbitrary real target activation gives 1.9\%, and a direction sampled from GCS~\citep{zhao2025} gives 5.3\%. The strongest single direction, namely, a tuned text-derived direction added to the source tokens~\citep{jiang2025}, yields 32.0\% (Table~\ref{tab:operators}). None of the single vectors, whether added or overwritten, exceeds this rate.

\begin{figure}[t]
    \centering
    \includegraphics[width=0.88\linewidth]{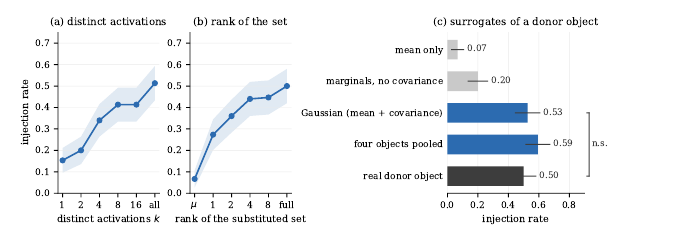}
    \caption{What insertion requires on the same 150 COCO images. (a) Increasing the number of distinct target activations raises injection while holding coverage fixed. (b) A low-rank reconstruction of the donor block shows that most of the effect appears by rank four. (c) The mean and marginal-preserving surrogates fail, whereas a Gaussian with the target first two population moments matches the real donor object. Pooling four objects performs better still. Shaded bands and error bars are 95\% binomial intervals.}
    \label{fig:requirements}
\end{figure}

The result changes when the substituted set contains multiple distinct activations. Figure~\ref{fig:requirements}(a) repeats $k$ target activations over the same $m$ positions while keeping coverage fixed. Injection rises monotonically from 0.15 for one activation to 0.51 for the full donor set. Figure~\ref{fig:requirements}(b) gives the same pattern when the donor block is reconstructed at increasing rank. Four directions already recover most of the effect.

The moment-matched surrogates in Table~\ref{tab:surrogates} identify the relevant property. The tiled mean and the marginal-preserving surrogate are both significantly below the real donor object. The Gaussian surrogate is statistically indistinguishable from it (11:15, $p=0.56$), even though it contains no real activation. Pooling activations from four objects exceeds the donor (10:24, $p=0.024$). These results support three conclusions. First, second-order structure is sufficient to match a real donor block within the resolution of this test, with a difference of $+0.027$ and a 95\% interval of $[-0.040,+0.093]$, whereas the individual numerical values of the block are not needed. Second, real target activations are unnecessary at edit time. Third, the joint structure of one particular object is unnecessary. Note that the marginals surrogate removes all dependence between dimensions and therefore does not, by itself, single out the covariance. We thus isolate the covariance within the Gaussian family in Section~\ref{sec:bank} and vary the higher-order structure in Appendix~\ref{app:nongauss}. The four-object result is consistent with a better class-level estimate giving stronger insertion. Section~\ref{sec:bank} tests this prediction with statistics estimated offline. The non-significant Gaussian-versus-donor test does not establish equality. With $n=150$, it excludes differences larger than approximately 0.1.

\begin{table}[t]
\caption{Moment-matched surrogates of a real donor object, $n=150$. Each condition is paired against the donor. The counts denote the images on which the donor succeeds while the surrogate fails, and conversely, and the last column reports the difference w.r.t.\ the donor with its 95\% interval. All rows are obtained from the same run. Bold and underline mark the best and second-best injection rates.}
\label{tab:surrogates}
\centering
\scriptsize
\setlength{\tabcolsep}{3.2pt}
\renewcommand{\arraystretch}{0.94}
\begin{tabular}{l l c c l l}
\toprule
surrogate & preserved & inject $\uparrow$ & retain $\uparrow$ & vs. donor & difference [95\% interval] \\
\midrule
mean & first moment & 0.067 & 0.536 & 66:1, $p=9\times10^{-19}$ & $-0.433$ $[-0.515,-0.352]$ \\
marginals & marginals & 0.200 & 0.513 & 49:4, $p=7\times10^{-11}$ & $-0.300$ $[-0.382,-0.218]$ \\
Gaussian & first and second moments & \underline{0.527} & 0.389 & 11:15, $p=0.56$ & $+0.027$ $[-0.040,+0.093]$ \\
four objects & class marginals & \textbf{0.593} & 0.368 & 10:24, $p=0.024$ & $+0.093$ $[+0.019,+0.168]$ \\
real donor object & all & 0.500 & 0.378 & -- & -- \\
\bottomrule
\end{tabular}
\end{table}

\subsection{Comparison with existing operators}
If the distributional account is correct, existing operators should rank according to the target second-order structure they transfer, even when they were designed for different purposes. AdaIN~\citep{huang2017adain} matches the per-channel mean and variance and therefore transfers the diagonal of the covariance. WCT~\citep{li2017wct} whitens source activations and recolors them with the target covariance. The steering operators of \citet{jiang2025}, \citet{li2025hidden}, and \citet{kravets2026} do not transfer target second-order structure. We restrict every operator to the object patches and tune its strength independently, as described in Appendix~\ref{app:baselines}.

Table~\ref{tab:operators} tests representative operators on both COCO and an independent LVIS benchmark. The LVIS evaluation uses 300 images from 50 classes that do not appear in the COCO evaluation, and every setting is frozen from COCO. The pooled Gaussian variants are strongest on both datasets and exceed a real donor object. WCT consistently exceeds AdaIN. The text-direction baseline is competitive on COCO but drops on LVIS and remains well below the Gaussian. The complete COCO baseline comparison, including GCS, residual steering, VISTA-SAE, WCT rank 16, retention, and paired tests, is reported in Appendix~\ref{app:baselines}, Table~\ref{tab:operators-full}. The source block spans a median of 96 out of 1024 dimensions, and WCT reaches the target covariance only up to a median relative error of 0.345. The pooled Gaussian is not limited by this source geometry.

\begin{table}[t]
\caption{Cross-dataset comparison of representative insertion operators. COCO uses the 150-image discovery set. LVIS uses 300 images from 50 classes absent from the COCO evaluation, with all settings frozen from COCO.}
\label{tab:operators}
\centering
\scriptsize
\setlength{\tabcolsep}{4.0pt}
\renewcommand{\arraystretch}{0.94}
\begin{tabular}{l c c c}
\toprule
operator & COCO inject $\uparrow$ & LVIS inject $\uparrow$ & LVIS retain $\uparrow$ \\
\midrule
mean only & 0.020 & 0.030 & 0.548 \\
text direction~\citep{jiang2025} & 0.320 & 0.283 & 0.546 \\
AdaIN~\citep{huang2017adain} & 0.180 & 0.190 & 0.558 \\
WCT, rank 4~\citep{li2017wct} & 0.307 & 0.320 & 0.502 \\
real donor object & 0.500 & 0.483 & 0.386 \\
ours, rank-4 Gaussian & 0.593 & \textbf{0.560} & 0.391 \\
ours, rank-8 Gaussian & \textbf{0.627} & \textbf{0.583} & 0.383 \\
\bottomrule
\end{tabular}
\end{table}

\subsection{Gaussian concept bank}
\label{sec:bank}
\begin{figure}[t]
    \centering
    \includegraphics[width=0.88\linewidth]{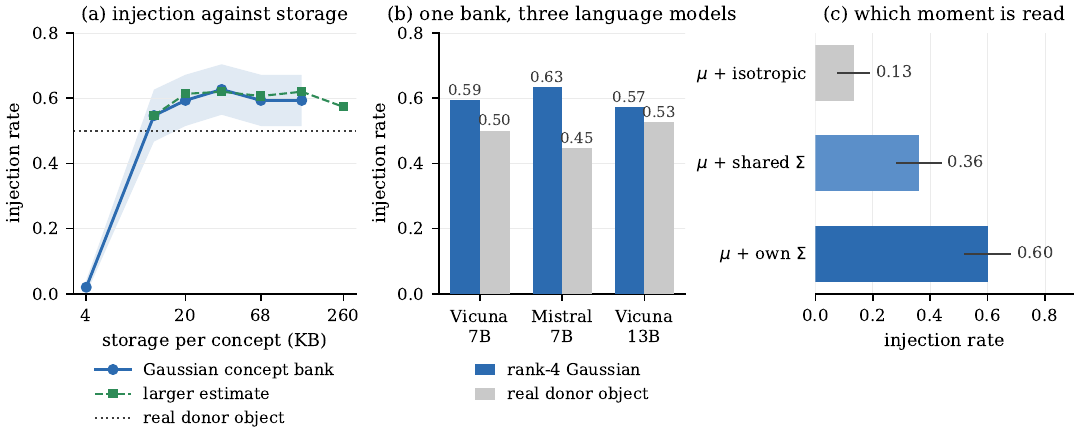}
    \caption{Gaussian concept bank. (a) Injection against storage per concept. The rank-4 representation uses 20~KB and exceeds a real donor object. The dashed curve repeats the sweep with statistics estimated from more images. (b) One CLIP-space bank is applied to three language models that share the vision encoder. (c) A shared covariance makes the mean readable, while the concept-specific covariance gives the strongest response and carries additional identity information.}
    \label{fig:bank}
\end{figure}

We next evaluate a bank estimated for 73 classes from a mean of 503 patches per class. Table~\ref{tab:storage} and Figure~\ref{fig:bank}(a) show how injection changes with rank. The mean alone gives 0.020, rank two gives 0.547, and rank four gives 0.593. Rank four exceeds a real donor object (28:14, $p=0.044$). Performance is approximately flat from rank four to rank 32. It declines at rank 64 when the bank is estimated from more images, which is consistent with the tail of the covariance estimate containing noise. We therefore use rank four as the compact CLIP-L/14 representation. Appendix~\ref{app:qual} gives qualitative examples.

We further study how many images are needed to estimate each concept while keeping the Gaussian rank fixed at four. With four estimation images, the rank-4 bank already reaches 0.560 injection, compared with 0.593 using twelve images, while retention remains nearly unchanged. This shows that most of the benefit is obtained from only a few examples, without increasing the 20~KB representation size. Appendix~\ref{app:heldout}, Figure~\ref{fig:sample-efficiency} and Table~\ref{tab:sample-efficiency}, give the full results.

The result is stable beyond the discovery set. On the 136 held-out images whose source is recognized, rank four gives 0.603 injection. On the 114 images whose source is not named by the unedited model, it gives 0.570. Across all 400 eligible COCO images, it gives 0.589. The donor object remains between 0.482 and 0.500 over these splits (Appendix~\ref{app:heldout}, Table~\ref{tab:heldout}). Insertion therefore does not depend on the model first naming the object that is replaced. Table~\ref{tab:operators} also reports the independent LVIS evaluation. With every setting frozen from COCO, the rank-4 Gaussian reaches 0.560 compared with 0.483 for the donor object (44:21, $p=5.9\times10^{-3}$), and the same operator ordering is preserved. Appendix~\ref{app:lvis}, Table~\ref{tab:lvisfreq}, breaks the LVIS result down by class frequency. The mean-versus-distribution gap also persists at other intervention points, across precisions and serving implementations, and with different projector architectures (Appendix~\ref{app:robust}, Tables~\ref{tab:robustsummary}--\ref{tab:projectors}).

\begin{table}[t]
\caption{Injection against storage cost, $n=150$, with each representation paired against a real donor object. The counts denote the images on which the representation succeeds while the donor fails, and conversely, and the last column reports the difference w.r.t.\ the donor with its 95\% interval. All rows are obtained from the same run, sampled independently of Table~\ref{tab:operators-full}.}
\label{tab:storage}
\centering
\scriptsize
\setlength{\tabcolsep}{3.0pt}
\renewcommand{\arraystretch}{0.94}
\begin{tabular}{l c c c l l}
\toprule
representation & storage $\downarrow$ & inject $\uparrow$ & full replacement $\uparrow$ & vs. donor & difference [95\% interval] \\
\midrule
mean only & 4 KB & 0.020 & 0.013 & 0:72, $p=4\times10^{-22}$ & $-0.480$ $[-0.560,-0.400]$ \\
mean, rank 2 & 12 KB & 0.547 & 0.400 & 26:19, $p=0.37$ & $+0.047$ $[-0.041,+0.134]$ \\
mean, rank 4 & 20 KB & \underline{0.593} & 0.433 & 28:14, $p=0.044$ & $+0.093$ $[+0.010,+0.177]$ \\
mean, rank 8 & 36 KB & \textbf{0.627} & 0.420 & 29:10, $p=3.4\times10^{-3}$ & $+0.127$ $[+0.048,+0.206]$ \\
mean, rank 16 & 68 KB & \underline{0.593} & \textbf{0.453} & 27:13, $p=0.039$ & $+0.093$ $[+0.012,+0.175]$ \\
real donor object & one image & 0.500 & 0.387 & -- & -- \\
\bottomrule
\end{tabular}
\end{table}

We then study whether the bank can be used with other language models. Vicuna-7B, Mistral-7B in the BakLLaVA configuration, and Vicuna-13B share the same CLIP feature space but use different projectors and language models. Their CLIP features are numerically identical up to a relative difference of $1.2\times10^{-2}$ and a mean cosine similarity of 0.9999. Since the bank is defined in the CLIP space, it can be applied to all three models without refitting. This experiment therefore evaluates whether three separately trained combinations of a projector and a language model all report an object when given synthetic tokens with four covariance directions, which is the case for each of them (Table~\ref{tab:transfer} and Figure~\ref{fig:bank}(b)). This shows that the reliance on low-rank second-order structure is not specific to one language model.

The mechanism also extends to different vision encoders, although the required rank changes. LLaVA-OneVision uses SigLIP-so400m-384 with 729 tokens of dimension 1152. The mean alone gives 0.047 and injection rises monotonically with rank. Rank four does not outperform the donor object (12:19, $p=0.28$), while rank eight reaches parity (20:15, $p=0.50$). The compression point is therefore a property of the representation and not a fixed constant of the method. On Qwen2.5-VL-7B, the mean alone gives 0.033 and rank four gives 0.300. Ranks eight and sixteen exceed the donor-object rate of 0.347 (23:10, $p=0.035$ and 26:8, $p=2.9\times10^{-3}$). The effect is therefore not specific to the LLaVA family.

\begin{table}[t]
\caption{Transfer across models. Values are injection rates for the stored Gaussian and the donor-object baseline of each model. Dashes denote ranks not evaluated in the fixed-rank transfer experiment.}
\label{tab:transfer}
\centering
\scriptsize
\setlength{\tabcolsep}{3.0pt}
\resizebox{0.92\textwidth}{!}{%
\begin{tabular}{l c c c c c}
\toprule
 & Vicuna-7B & Mistral-7B & Vicuna-13B & OneVision & Qwen2.5-VL \\
vision encoder & \multicolumn{3}{c}{CLIP-L/14-336} & SigLIP-384 & Qwen ViT \\
\midrule
mean only & 0.020 & 0.020 & 0.033 & 0.047 & 0.033 \\
rank-4 Gaussian & 0.593 & \textbf{0.633} & \textbf{0.573} & 0.300 & 0.300 \\
rank-8 Gaussian & \textbf{0.627} & -- & -- & 0.380 & 0.433 \\
rank-16 Gaussian & 0.593 & -- & -- & \textbf{0.387} & \textbf{0.467} \\
real donor object & 0.500 & 0.447 & 0.527 & 0.347 & 0.347 \\
\bottomrule
\end{tabular}}
\end{table}

Finally, we vary the first and second moments independently to determine which one carries concept identity. A mean with isotropic noise of matched total variance gives 0.133. The same mean with a pooled within-class covariance shared across all 73 concepts gives 0.360 (using all 64 stored directions). Using the covariance of the concept itself raises injection to 0.600 (using 32 directions). Note that none of these sets has structure beyond the second order, and the differences among them are thus due to the covariance. Thus, covariance first makes the mean readable and then contributes concept-specific information. When the mean and covariance come from different concepts, the covariance concept is named more often than the mean concept (0.353 versus 0.227, 44:25, $p=0.030$). Reversing the assignment gives the same direction but is not significant (37:28, $p=0.32$). Subtracting each class's rate in the unedited captions changes these differences by less than 0.001. A compact bank that shares one covariance across all classes, the pooled within-class covariance truncated to rank eight, gives 0.440 while storing only the class mean at 4~KB per concept. This is indistinguishable from a real donor object (33:24, $p=0.29$), but remains below the per-class covariance (31:7, $p=1.2\times10^{-4}$). Furthermore, this gap is not caused by the rank. The shared covariance yields 0.400, 0.447, 0.440, 0.433, and 0.387 at ranks 2, 4, 8, 32, and 64, respectively, whereas the per-class covariance yields 0.547, 0.593, 0.627, and 0.593 at ranks 2, 4, 8, and 32 (Appendix~\ref{app:nongauss}).

\subsection{Direct questioning}
\label{sec:direct}
A caption-level string match could reflect lexical emission without a broader change in the model's answers. We therefore ask closed questions about object presence. Before editing, the model answers ``no'' for the absent target in 98.0\% of cases and ``yes'' for the source object in 98.7\%. After substituting the rank-4 Gaussian, it answers ``yes'' for the target in 88.0\%, corresponding to a change on 129 of 150 images. Its affirmation of the source drops to 50.0\%.

We next test specificity. Under the same intervention, we ask whether a third, unrelated concept is present. The model affirms this control in 5.3\% of edited images, compared with 2.7\% before editing. The target is affirmed while the control is rejected on 122 images, and the reverse never occurs ($p=3.8\times10^{-37}$). We then repeat the test with two prompts that do not presuppose the target's presence. Table~\ref{tab:prompts} reports all three prompt conditions. Free recall still names the target in 0.647 of edited images while naming the unrelated control in only 0.033. Forced choice selects the target in 0.800 of cases and the control in 0.027. The effect is therefore robust to prompt format and remains specific to the inserted concept. Appendix~\ref{app:questions} gives the exact prompt text and the concept-level susceptibility analysis.

\begin{table}[t]
\caption{Direct questioning on the same 150 images and rank-4 Gaussian edit. The target is the inserted concept and the control is an absent, unrelated concept.}
\label{tab:prompts}
\centering
\scriptsize
\setlength{\tabcolsep}{2.4pt}
\renewcommand{\arraystretch}{0.92}
\resizebox{0.92\textwidth}{!}{%
\begin{tabular}{l c c c}
\toprule
 & yes/no & inventory & forced choice \\
\midrule
target affirmed, unedited image $\downarrow$ & 0.020 & 0.013 & 0.020 \\
target affirmed, edited image $\uparrow$ & 0.880 & 0.647 & 0.800 \\
control affirmed, edited image $\downarrow$ & 0.053 & 0.033 & 0.027 \\
source affirmed, edited image & 0.500 & 0.460 & 0.533 \\
target vs. control, edited & 122:0, $p=3.8\times10^{-37}$ & 94:2, $p=1.2\times10^{-25}$ & 117:1, $p=7.2\times10^{-34}$ \\
\bottomrule
\end{tabular}}
\end{table}

The direct questions also show that caption matching is conservative for this condition. The caption omits the target while the model affirms it on 49 images. The reverse occurs on only six ($p=1.8\times10^{-9}$). Concept susceptibility is not uniform. Discrete objects of intermediate size are affirmed reliably, while thin implements and large furniture are less susceptible. Appendix~\ref{app:questions} reports the concept-level profile.

\subsection{Removal}
\label{sec:removal}
In this section, we study whether a single direction suffices to remove an object and whether this removal is selective. In contrast to insertion, no token is substituted. Instead, a direction is projected out of the existing tokens in $P$, i.e., the set of tokens covering the object. Removal uses a separate clustered dictionary of approximately $K=230$ visual regions. We build it from a held-out activation stream using a cosine threshold. For a source object, we identify the region whose centroid is most over-represented inside $P$ relative to the rest of the image, and project that region direction out of the object tokens with strength $\beta$. At the strength used in the fidelity sweep, this removes the object from 39.3\% of captions. A random direction at matched norm removes it in 0.7\%. The model is therefore insensitive to generic perturbations of the same magnitude.

\begin{figure}[t]
    \centering
    \includegraphics[width=0.88\linewidth]{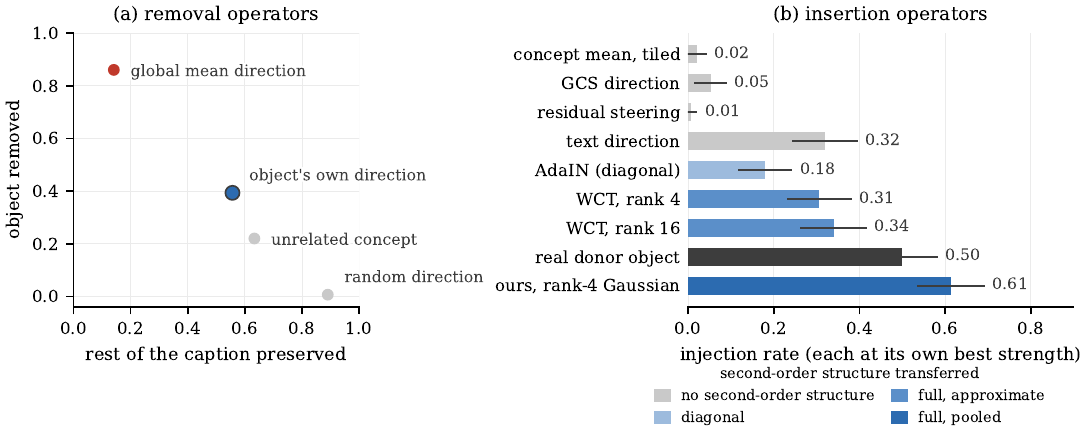}
    \caption{Removal and insertion are not symmetric. (a) Removal operators are plotted by the fraction of captions in which the source disappears against the fraction of the remaining caption that is preserved. A single object direction can remove the source while retaining most of the scene description. (b) Insertion operators on the same images rank with the target second-order structure they transfer. Operators that transfer no target covariance perform worst, AdaIN transfers only the diagonal, WCT approximates full covariance, and the pooled Gaussian performs best.}
    \label{fig:asymmetry}
\end{figure}

Figure~\ref{fig:asymmetry}(a) shows why removal cannot be judged by removal rate alone. Projecting out the global mean direction removes the object in 86.0\% of cases but preserves only 14.1\% of the remaining caption. The object's region direction gives the best balance among the four operators. This rank-one removal result is consistent with linear concept-erasure theory. \citet{belrose2023} show that linear guarding can be removed by equalizing class-conditional means and eliminating feature-label cross-covariance. \citet{avitan2026} further show that within-class covariance asymmetry remains after mean equalization and can still support nonlinear recovery. Our insertion results provide a complementary causal observation: the model acts on the second-order structure that a first-moment intervention leaves behind.

At its tuned strength $\beta=20$, our operator removes the source in 44.7\% of cases. A tuned text-derived direction reaches 54.7\% (29:14, $p=0.031$). Our operator is more stable as the strength increases. Over $\beta\in[1,160]$, its retention never falls below 0.47, whereas the text-direction baseline drops from 0.414 at $\beta=20$ to 0.050 at $\beta=40$ (Appendix~\ref{app:removal}).

\paragraph{Why removal and insertion differ.}
Removal requires disrupting the representation of the object, which is achieved even by operators that fail to insert. For instance, the tiled mean inserts its target in 6.7\% of the captions but removes the source in 44.0\%. By contrast, insertion requires producing tokens that the language model interprets as an object, which depends on the variation among these tokens. In particular, a block of identical tokens yields 0.020, unstructured variation yields 0.133, variation shared across objects yields 0.360 to 0.447, and object-specific variation yields 0.593 to 0.627. Consequently, the best rank-one edit (0.320) reaches approximately half of the rate obtained with second-order structure on the same images (53:9, $p=1.1\times10^{-8}$).

\subsection{Insertion without annotation}
The experiments above use the ground-truth box to select intervention positions. Insertion is much less dependent on exact location. A centered square with the same area as the object gives 0.520 injection, compared with 0.587 for the annotated box (23:13, $p=0.13$). A fixed $8\times8$ centered square, which uses no scene information, gives 0.460 with a retention of 0.462. A randomly placed square of matched area gives 0.473 with a retention of 0.470. At a comparable retention, WCT and VISTA-SAE yield injection rates between 0.273 and 0.340 (Appendix~\ref{app:fidelity}). Insertion is therefore sensitive to how much of the visual stream is replaced but comparatively insensitive to the precise location. Full replacement behaves differently. It falls from 0.420 with the annotated box to 0.173 with a centered square because removing the source does require its location. Replacing all 576 tokens yields 0.680 injection but only 0.019 retention, at which point the model describes almost exclusively the inserted concept.

\section{Conclusion and Limitations}
We have shown that removal and insertion behave differently in frozen VLM visual-token streams. A rank-one modification can suppress an object, whereas insertion depends on second-order structure: the covariance makes the concept mean readable, and the concept-specific covariance yields the highest insertion rates. A real donor is unnecessary. On CLIP-L/14, its distributional estimate is stored at rank four in 20~KB per concept. SigLIP and Qwen require modestly higher rank, so the compression point depends on the representation. The mechanism replicates on an independent long-tailed benchmark, at multiple visual-to-language intervention points, and across precisions, serving implementations, and projector architectures. Experiments are limited to object concepts and white-box access. Furthermore, replacing the object tokens changes the remaining caption to the same extent as transplanting a real object, and a larger part of the scene is preserved when fewer tokens are replaced, at a lower insertion rate (Appendix~\ref{app:fidelity}). Within this scope, low-dimensional second-order structure supports training-free counterfactual interventions, and systems that expose intermediate activations should protect the visual-to-language activation path.

\section*{Ethics Statement}
This work studies inference-time manipulation of a frozen VLM by modifying its internal visual activations without changing the image or model weights. The intervention requires access to activations passed between the vision encoder and the language model. It is therefore a white-box attack on the inference pipeline and not an image-space attack. A relevant threat model is a compromised serving component or an untrusted adapter that can modify this activation stream. An adversary limited to supplying input images cannot perform the intervention studied here. Once internal access is available, the procedure requires only stored concept statistics and no gradient computation. Protecting the integrity of activations exchanged between model components is therefore a natural mitigation. We report this behavior to support evaluation of systems that expose or modify intermediate representations. All model experiments use public models and the public COCO and LVIS datasets. Two annotators evaluated model-generated captions (Appendix~\ref{app:human}). No personal, sensitive, or private participant data were collected.

\appendix

\section{Protocol Details}
\label{app:protocol}
A forward pre-hook on the multimodal projector captures the $576\times1024$ visual-token matrix and replaces the selected rows. Each substitute is scaled to the norm of the activation it replaces.

We construct the COCO evaluation set from images in which a single instance of a source class occupies between 2\% and 30\% of the frame. This gives 400 images over 73 classes. An image is scorable for removal when the unedited caption contains the class name or a listed synonym, which holds for 286 images. The condition sweeps use the first 150 scorable images in a fixed order.

For an image $i$ with source class $c$, the target is selected as the $i$-th element, modulo the list length, from the sorted list of classes other than $c$. The same assignment is used for every condition and experiment.

Injection and removal are scored with word-boundary regular-expression matches against the lower-cased class name and its synonyms. Retention is the Jaccard-style proportion of content words from the unedited caption that remain in the edited caption after stop words and both class names are excluded.

For each concept, we pool activations inside the bounding boxes of up to twelve images and store the mean and the leading 32 covariance eigenpairs. The pooled within-class covariance is the sum of per-class scatter matrices divided by the total degrees of freedom, and we store its leading 64 eigenpairs. The surrogate constructions are given in Section~3.2. Gaussian mixing coefficients are drawn once per image from a seed derived from the image identifier, making the surrogate deterministic across conditions.

All COCO images are taken from val2017, and each of the 400 evaluation images has one designated object. The statistics of a class are pooled from the boxes of the first images of this pool whose designated object belongs to that class, and the donor objects are drawn from the first four of these images. In particular, the bank of Section~\ref{sec:bank} uses up to twelve images per class, whereas the operator comparison of Table~\ref{tab:operators-full} and the controls of Appendices~\ref{app:norm} to~\ref{app:fidelity} use the first four. Since an image with source class $c$ is edited toward a target $t\neq c$, no image is edited with statistics or a donor obtained from itself, which we verified for all 150 discovery images. Furthermore, the bank consists of a mean and a covariance of encoder activations and is thus not fitted to any outcome. The rank, which is the single selected setting, was chosen on the 150 discovery images and is used unchanged on the held-out images, on the unrecognized images, and on LVIS, whose statistics are estimated from the LVIS training split.

All experiments run on a single NVIDIA RTX 5000 Ada or RTX A5000 GPU. Three models use four-bit precision. LLaVA-OneVision uses half precision because its MLP width of 4304 is not aligned with the block size of the four-bit kernel. Appendix~\ref{app:robust} verifies that this precision difference does not change the mechanism.

\section{Baseline Tuning}
\label{app:baselines}
Every baseline has a strength parameter, so we tune each method independently and do not evaluate it at the strength used for our operator. We report the best injection rate subject to retention of at least 0.35. Text-direction insertion is evaluated at $\alpha\in\{0.25,0.5,1,2,4\}$ and reported at $\alpha=1$. Text-direction removal is evaluated at $\beta\in\{1,2,5,10,20,40\}$ and reported at $\beta=20$. Residual-stream steering is evaluated at the same $\alpha$ values. No setting both inserts and retains, so we report $\alpha=0.25$. Our removal operator is reported at $\beta=20$.

These baselines re-implement the published intervention primitives and restrict them to the object patches. They are not the complete released systems. VISTA-SAE~\citep{kravets2026} is the only baseline that requires training. We use the authors' released projector for LLaVA-1.5-7B and apply its replacement operator at the object patches using the SAE feature of the target. We sweep $\{0.25,0.5,1,2,4\}$ and report strength 1.0. VISTA-SAE reaches 0.273 injection at 0.482 retention. This exceeds the other single-vector operators but remains well below the Gaussian (53:2, $p=8.6\times10^{-14}$), consistent with the account in Section~\ref{sec:requirements}. Table~\ref{tab:operators-full} gives the complete COCO comparison underlying the representative cross-dataset results in Table~\ref{tab:operators}.

\begin{table}[h]
\caption{Complete COCO operator comparison on the 150-image discovery set. Each method is evaluated at its best strength and paired against our operator. The counts denote the images on which our operator succeeds while the other fails, and conversely. Table~\ref{tab:paired} reports the run of each row and the complete paired tables. $\dagger$ indicates a method that trains a projector.}
\label{tab:operators-full}
\centering
\scriptsize
\setlength{\tabcolsep}{2.6pt}
\resizebox{\textwidth}{!}{%
\begin{tabular}{l l c c l}
\toprule
operator & target second-order structure & inject $\uparrow$ & retain $\uparrow$ & vs. ours \\
\midrule
concept mean, tiled & none & 0.020 & 0.557 & 89:0, $p=3\times10^{-27}$ \\
GCS direction~\citep{zhao2025} & none & 0.053 & 0.508 & 85:1, $p=2\times10^{-24}$ \\
residual steering~\citep{li2025hidden} & none & 0.007 & 0.798 & 91:0, $p=8\times10^{-28}$ \\
text direction~\citep{jiang2025} & none & 0.320 & 0.535 & 53:9, $p=1.1\times10^{-8}$ \\
VISTA-SAE$^\dagger$~\citep{kravets2026} & none & 0.273 & 0.482 & 53:2, $p=8.6\times10^{-14}$ \\
AdaIN~\citep{huang2017adain} & diagonal & 0.180 & 0.570 & 65:0, $p=5\times10^{-20}$ \\
WCT, rank 4~\citep{li2017wct} & full, approximate & 0.307 & 0.490 & 47:1, $p=3\times10^{-13}$ \\
WCT, rank 16~\citep{li2017wct} & full, approximate & 0.340 & 0.479 & 42:1, $p=1\times10^{-11}$ \\
real donor object & full, one sample & 0.500 & 0.378 & 28:11, $p=9.5\times10^{-3}$ \\
ours, rank-4 Gaussian & full, pooled & \textbf{0.613} & 0.376 & -- \\
\bottomrule
\end{tabular}}
\end{table}

\section{Removal Strength}
\label{app:removal}
\begin{figure}[h]
    \centering
    \includegraphics[width=0.58\linewidth]{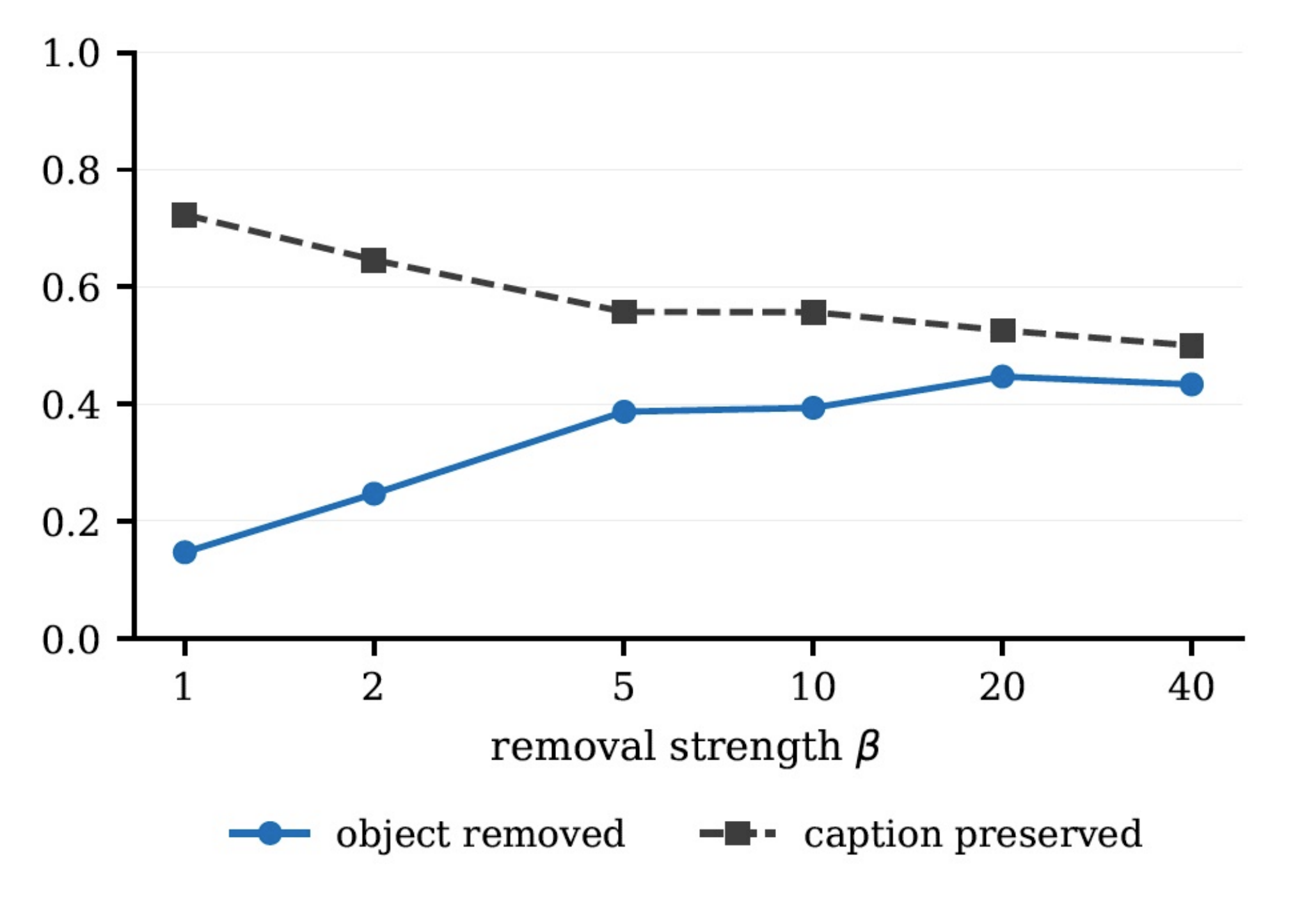}
    \caption{Removal as a function of intervention strength $\beta$. The solid line is the fraction of captions in which the object disappears, and the dashed line is the fraction of the remaining caption content that is preserved. Removal saturates near $\beta=20$, while retention degrades gradually. The tuned text-direction baseline reaches a higher peak removal rate, but its retention falls from 0.41 at $\beta=20$ to 0.05 at $\beta=40$.}
    \label{fig:removal-strength}
\end{figure}

Figure~\ref{fig:removal-strength} shows the complete strength sweep for our removal operator and the tuned text-direction baseline. The object-removal rate saturates near $\beta=20$, whereas retention degrades more gradually for our operator.

\section{Additional Controls and Negative Results}
\label{app:controls}
We first test whether the spatial arrangement of substitute activations matters. Permuting them within the object changes the outcome on only 17 of 286 images (17:14, $p=0.72$). The same null result holds for Mistral-7B (11:16, $p=0.44$) and Vicuna-13B (9:15, $p=0.31$). On the same three models, the full donor set strongly outperforms its tiled mean, with discordant counts of 124:1 ($p=6\times10^{-36}$), 97:1 ($p=6\times10^{-28}$), and 121:2 ($p=1\times10^{-33}$), respectively, for $n\geq259$. This agrees with the observation of \citet{schaumloffel2026} that token arrangement inside an object boundary has little effect on reading out a bounding box. We observe the same behavior when writing one in.

The clustered dictionary used for removal partitions the visual stream into $K$ regions. Across three complete rebuilds, $K$ ranges from 227 to 245. Operators that do not use this dictionary have an injection standard deviation of 0.000 across rebuilds. Operators that do use it have standard deviations of 0.016 and 0.017.

\begin{figure}[h]
    \centering
    \includegraphics[width=0.52\linewidth]{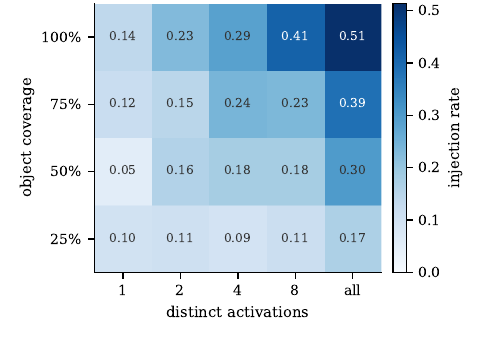}
    \caption{Injection as a function of the proportion of object tokens replaced and the number of distinct donor activations spread over those positions, $n=150$. Neither coverage nor diversity is sufficient alone. Increasing diversity changes injection by only 0.07 at 25\% coverage but by 0.37 at full coverage.}
    \label{fig:coverage}
\end{figure}

Coverage and activation diversity are both necessary. At 25\% object coverage, increasing diversity changes injection by 0.067. At full coverage, the same change is 0.373. Injection is approximately linear in coverage, and the absolute box size is only weakly correlated with injection (0.175). Figure~\ref{fig:coverage} shows the full grid.

Across four donor images per class, injection is $0.512\pm0.030$. At least one donor succeeds on 78.0\% of images, while all four succeed on 22.0\%.

We also tested two diagnostics that did not support useful conclusions. First, we asked for the color of an inserted object in an attempt to distinguish a perceptual change from lexical emission. All 132 valid edited responses named a color. However, all 150 responses also named a color when asked about a control concept that the model had explicitly denied was present. The model accepts the presupposition of the question, so this probe does not discriminate the mechanisms of interest. Second, we predicted that the rank at which injection saturates would correlate with the participation-ratio dimension of the donor covariance spectrum. The per-image correlation is $\rho=-0.071$ ($p=0.51$, $n=86$). We also predicted a coverage threshold, but a logistic fit does not yield an interpretable midpoint.

\section{Direct Questioning: Prompts and Susceptibility}
\label{app:questions}
Section~\ref{sec:direct} uses the yes/no question \emph{Is there a \{target\} in this image? Answer yes or no.} We repeat the experiment on the same 150 images and edits with two additional prompts. The inventory prompt, \emph{List every object you can see in this image, separated by commas. Do not describe the scene}, never names the target. A target in the answer must therefore come from the model. We score the target in the list as injection and the unrelated control concept as a specificity error. The forced-choice prompt asks, \emph{Which of these is in the image: a \{source\}, a \{target\}, both, or neither? Answer with one word.} This prompt offers both concepts together with a ``neither'' option and therefore measures removal, insertion, and replacement without presupposing that either object is present.

Table~\ref{tab:prompts} in Section~\ref{sec:direct} compares the three prompts. Free recall is lower than yes/no recognition (0.647 versus 0.880), as expected, but remains far above the unrelated control (0.033). Forced choice also strongly favors the target, with 0.800 target selection and 0.027 control selection. The effect therefore persists without presupposing the target's presence.

Susceptibility varies by concept. Airplane, apple, backpack, banana, baseball bat, and bear are affirmed on every image on which they are inserted. The least susceptible concepts are fork (0.33), chair (0.00), dining table (0.00), and spoon (0.00). Thin implements occupy few patches, while large furniture can have token statistics that overlap with the background. Both explanations are consistent with the coverage experiment, but we do not test them separately.

\section{Replication on Held-Out and Out-of-Selection Images}
\label{app:heldout}
The 400 images satisfying the size criterion are split according to whether the unedited model names the source. Removal can only be scored on the 286 recognized images, so the mechanistic experiments use the first 150 for discovery and the remaining 136 as a held-out set. Insertion can be scored on any image since its scoring does not require the unedited caption to name the source. We therefore also run the final operator on the 114 images whose source is not named, using the same target-assignment procedure. Table~\ref{tab:heldout} reports the three splits and their union.

\begin{table}[h]
\caption{Injection of the rank-4 Gaussian on the discovery set, held-out recognized images, images whose source is not named by the unedited model, and their union.}
\label{tab:heldout}
\centering
\small
\begin{tabular}{l r c c c}
\toprule
split & $n$ & mean only $\uparrow$ & rank-4 Gaussian $\uparrow$ & real donor object $\uparrow$ \\
\midrule
discovery, source recognized & 150 & 0.020 & 0.593 & 0.500 \\
held out, source recognized & 136 & 0.022 & 0.603 & 0.493 \\
source not recognized & 114 & 0.026 & 0.570 & 0.482 \\
all eligible images & 400 & 0.023 & 0.589 & 0.493 \\
\bottomrule
\end{tabular}
\end{table}

We also vary the number of images used to estimate each concept while keeping the Gaussian rank fixed at four. Table~\ref{tab:sample-efficiency} reports this sample-efficiency experiment. The stored representation remains 20~KB per concept for every setting.

\begin{figure}[t]
    \centering
    \includegraphics[width=0.56\linewidth]{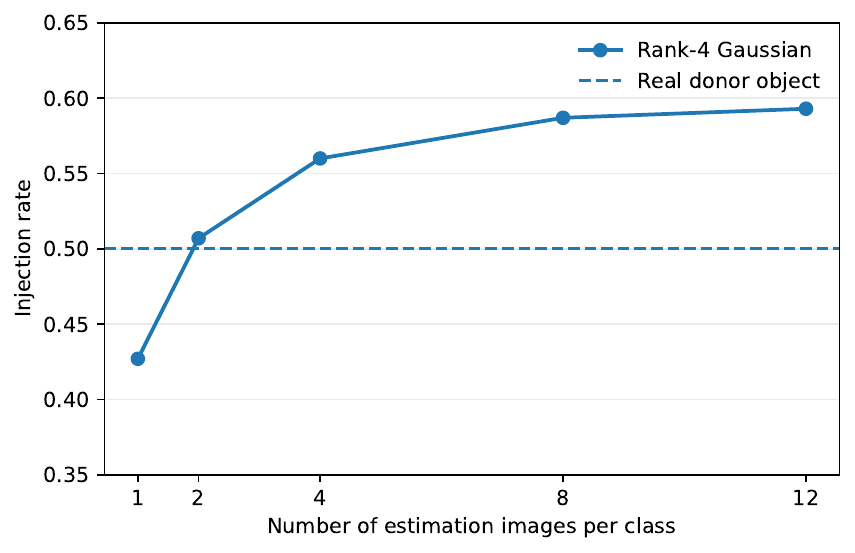}
    \caption{Sample efficiency of the rank-4 concept bank. Injection improves rapidly as the number of images used to estimate each concept increases, reaching 0.560 with four images and 0.593 with twelve. The dashed line shows the real-donor baseline at 0.500. Retention remains nearly constant over the sweep, changing from 0.385 to 0.376.}
    \label{fig:sample-efficiency}
\end{figure}

\begin{table}[h]
\caption{Sample efficiency of the rank-4 Gaussian concept bank on the 150-image discovery set. Storage is unchanged as the number of estimation images increases.}
\label{tab:sample-efficiency}
\centering
\small
\setlength{\tabcolsep}{5pt}
\begin{tabular}{l c c c c}
\toprule
estimation images / class & rank & storage / concept & inject $\uparrow$ & retain $\uparrow$ \\
\midrule
1 & 4 & 20 KB & 0.427 & 0.385 \\
2 & 4 & 20 KB & 0.507 & 0.381 \\
4 & 4 & 20 KB & 0.560 & 0.379 \\
8 & 4 & 20 KB & 0.587 & 0.377 \\
12 & 4 & 20 KB & \textbf{0.593} & 0.376 \\
real donor object & -- & one image & 0.500 & 0.378 \\
\bottomrule
\end{tabular}
\end{table}

\section{Independent Benchmark: LVIS}
\label{app:lvis}
We evaluate the mechanism on LVIS-v1~\citep{gupta2019} to move beyond the 73-class COCO vocabulary. LVIS provides a broader, more fine-grained, and long-tailed set of object categories. We use 300 validation images from 50 classes that do not appear in the COCO evaluation, with six images per class. Concept statistics are estimated independently from up to twelve LVIS training images per class. Targets are absent according to the LVIS annotations. Every setting, including the rank, is frozen from COCO and no hyperparameter is tuned on LVIS.

Table~\ref{tab:operators} in Section~\ref{sec:bank} compares the representative operators on COCO and LVIS. The ordering from Section~4.3 replicates, and the rank-4 Gaussian gives 0.560 injection compared with 0.483 for a donor object (44:21, $p=5.9\times10^{-3}$). Injection declines from frequent to rare classes, as shown in Table~\ref{tab:lvisfreq}. This indicates that estimate quality still matters in the long tail. Nevertheless, the pooled Gaussian remains above a real donor object in every frequency stratum, and the mean-versus-covariance effect is not specific to COCO.

\begin{table}[h]
\caption{LVIS injection by frequency stratum, with 100 images in each stratum.}
\label{tab:lvisfreq}
\centering
\small
\begin{tabular}{l r c c c}
\toprule
stratum & $n$ & mean only & real donor object & rank-4 Gaussian \\
\midrule
frequent & 100 & 0.030 & 0.500 & 0.580 \\
common & 100 & 0.030 & 0.490 & 0.570 \\
rare & 100 & 0.030 & 0.460 & 0.530 \\
all & 300 & 0.030 & 0.483 & 0.560 \\
\bottomrule
\end{tabular}
\end{table}

\section{Robustness of the Distributional Mechanism}
\label{app:robust}
The main experiment intervenes at the projector input of LLaVA-1.5-7B, uses four-bit precision, runs through the Hugging Face eager implementation, and uses the model's two-layer MLP projector. We vary these choices one at a time on the same 150 images and target assignments. Table~\ref{tab:robustsummary} summarizes the results, and Tables~\ref{tab:points}--\ref{tab:projectors} give the detailed comparisons. In every setting, the tiled mean remains between 0.02 and 0.04 while the Gaussian remains between 0.51 and 0.60. Absolute rates change, but the gap between a mean and a distribution persists.

We first move the intervention point. We intervene at an earlier vision-tower layer, at the projector output after conversion to the language-model embedding space, and in the language-model residual stream after the first transformer block. For each location, we estimate $\mu_c$ and $\Sigma_c$ directly in that representation so the test is not confounded by a coordinate mismatch. At rank eight, the Gaussian exceeds the donor object at all four intervention points. The paired counts from earliest to latest are 29:13 ($p=0.020$), 29:10 ($p=3\times10^{-3}$), 25:10 ($p=0.017$), and 22:8 ($p=0.016$). The absolute rate declines later in the pathway, but the need for a distribution is not specific to the encoder-projector boundary.

We next vary numerical precision and serving implementation. Table~\ref{tab:serving} reports these comparisons. With the same weights, half and four-bit precision yield 0.600 and 0.593 injection. The Gaussian outcome differs on only nine of 150 images (4:5, $p=1.0$). Across Hugging Face eager attention, FlashAttention-2, and an independently instrumented vLLM serving path, binary outcomes agree on at least 96\% of images and no pairwise difference is significant ($p>0.5$).

Finally, we vary the projector while holding the vision encoder, language model, and training procedure fixed. Linear, two-layer MLP, and gated MLP projectors all preserve the same qualitative gap. The mean gives 0.02--0.03 injection and the Gaussian gives 0.56--0.59. The second-order requirement therefore originates upstream of the projector architecture.

\begin{table}[h]
\caption{Robustness summary on $n=150$ images.}
\label{tab:robustsummary}
\centering
\small
\begin{tabular}{l c c}
\toprule
setting & rank-4 Gaussian $\uparrow$ & mean only $\uparrow$ \\
\midrule
projector input (main setting) & 0.593 & 0.020 \\
earlier vision-tower layer & 0.553 & 0.027 \\
projector output & 0.547 & 0.033 \\
language-model residual, block 1 & 0.507 & 0.040 \\
half precision & 0.600 & 0.020 \\
four-bit precision & 0.593 & 0.020 \\
FlashAttention-2 & 0.587 & 0.020 \\
vLLM serving path & 0.580 & 0.027 \\
linear projector & 0.560 & 0.027 \\
gated MLP projector & 0.573 & 0.027 \\
\bottomrule
\end{tabular}

\caption{Injection at each intervention point, $n=150$, with concept statistics estimated directly in each representation.}
\label{tab:points}
\centering
\small
\resizebox{\textwidth}{!}{%
\begin{tabular}{l c c c c}
\toprule
intervention point & mean only & rank-4 Gaussian & rank-8 Gaussian & real donor object \\
\midrule
earlier vision-tower layer & 0.027 & 0.553 & 0.580 & 0.480 \\
projector input (main setting) & 0.020 & 0.593 & 0.627 & 0.500 \\
projector output & 0.033 & 0.547 & 0.580 & 0.460 \\
language-model residual, block 1 & 0.040 & 0.507 & 0.553 & 0.433 \\
\bottomrule
\end{tabular}}

\caption{Precision and serving implementation for the same LLaVA-1.5-7B weights, $n=150$.}
\label{tab:serving}
\centering
\small
\begin{tabular}{l c c c c}
\toprule
setting & mean only & rank-4 Gaussian & real donor object & retention \\
\midrule
half precision, eager & 0.020 & 0.600 & 0.500 & 0.381 \\
four-bit, eager (main) & 0.020 & 0.593 & 0.493 & 0.376 \\
FlashAttention-2 & 0.020 & 0.587 & 0.493 & -- \\
vLLM serving path & 0.027 & 0.580 & 0.493 & -- \\
\bottomrule
\end{tabular}

\caption{Projector architecture with vision encoder, language model, and training procedure fixed, $n=150$.}
\label{tab:projectors}
\centering
\small
\begin{tabular}{l c c c c}
\toprule
projector & mean only & rank-4 Gaussian & rank-8 Gaussian & real donor object \\
\midrule
linear & 0.027 & 0.560 & 0.587 & 0.480 \\
two-layer MLP (main) & 0.020 & 0.593 & 0.627 & 0.500 \\
gated MLP & 0.027 & 0.573 & 0.600 & 0.487 \\
\bottomrule
\end{tabular}
\end{table}

\section{Human Validation of the Caption Metric}
\label{app:human}
Two annotators who were blind to the condition independently evaluated the 150 edited captions from the rank-4 Gaussian condition. For each caption, they judged whether it described the target concept as present in the scene, which corresponds to injection. Inter-annotator agreement is 99.3\% with Cohen's $\kappa=0.99$. The annotators resolved their disagreement by consensus before comparison with the automatic metric.

Metric and human judgment agree on 148 of 150 captions (98.7\%, $\kappa=0.97$). Table~\ref{tab:human} gives the corresponding confusion matrix. Both disagreements are false negatives of the automatic metric. It never credits a caption that humans judge as not describing the target, but it misses two descriptions that avoid the literal class name: a baseball player holding a bat for \emph{baseball bat}, and a glass of wine for \emph{wine glass}. The automatic injection metric is therefore conservative, consistent with the direct-question comparison in Section~\ref{sec:direct}.

\begin{table}[h]
\caption{Automatic injection score against human judgment on 150 edited captions. Agreement is 98.7\% with Cohen's $\kappa=0.97$.}
\label{tab:human}
\centering
\small
\begin{tabular}{l c c}
\toprule
 & human: target present & human: target absent \\
\midrule
metric: target named & 89 & 0 \\
metric: target not named & 2 & 59 \\
\bottomrule
\end{tabular}
\end{table}

\section{Qualitative Examples}
\label{app:qual}
\begin{figure}[h]
    \centering
    \includegraphics[width=\linewidth]{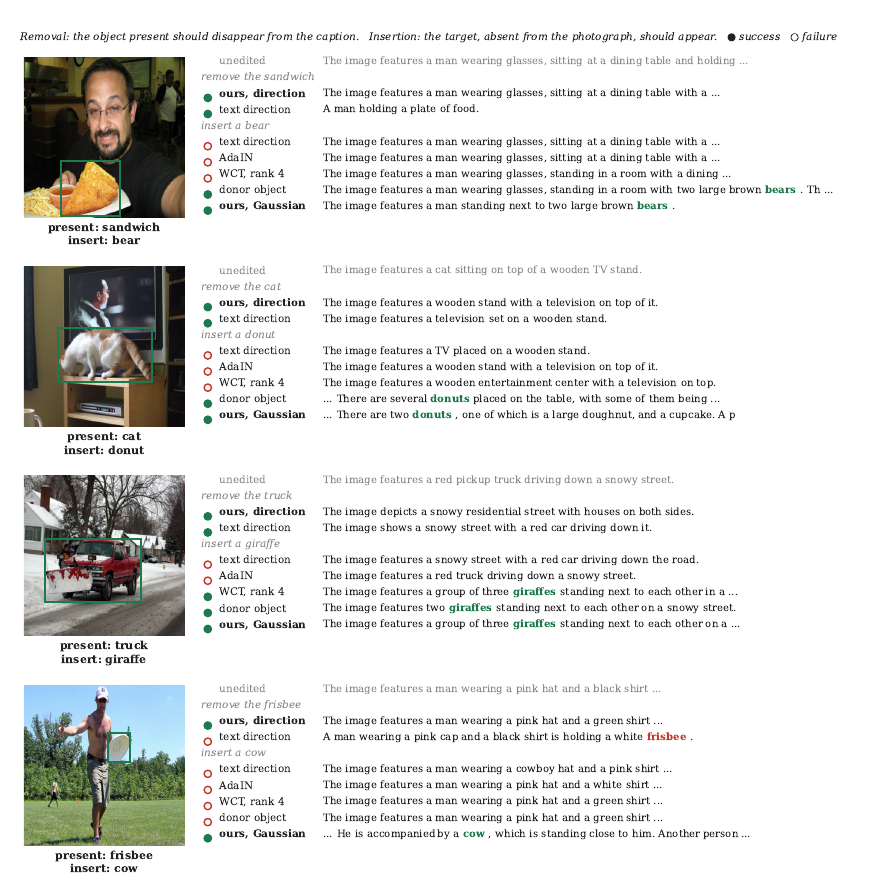}
    \caption{Qualitative comparison of removal and insertion on the same images. Each row shows one COCO image, the source object, and an absent target concept. Filled markers denote success. The removal block compares the object's clustered-dictionary direction with a text-derived direction. The insertion block compares a text direction, AdaIN, WCT, a real donor object, and the rank-4 Gaussian. The Gaussian succeeds on all four examples.}
    \label{fig:qual}
\end{figure}

Figure~\ref{fig:qual} compares the operators from Table~\ref{tab:operators} on representative images. The target concept is absent from every photograph. A successful insertion therefore requires the edited caption to report a concept that is not in the input image. Removal can succeed with a single direction. Among the insertion methods shown, the Gaussian is the only operator that succeeds on all four examples.

\clearpage
\raggedbottom
\section{Paired Tables for the Operator Comparison}
\label{app:paired}
In Table~\ref{tab:paired}, we report the complete paired tables w.r.t.\ the rank-4 Gaussian of the operator-comparison run, which succeeds on 92 of the 150 images. The rates of the text direction and of residual steering were measured in the baseline-tuning run, whereas that of VISTA-SAE was measured in a separate run. Note that all runs use the same 150 images in the same order with the same target for each image, and each row is thus paired image by image. In each row, the number of successes of the operator is the sum of the fourth and sixth columns, and the fourth and fifth columns sum to 92.

\begin{table}[H]
\caption{Paired tables underlying Table~\ref{tab:operators-full} ($n=150$). Our operator is the rank-4 Gaussian, which succeeds on 92 images. A row favors our operator when ``ours only'' exceeds ``other only''. The last column reports the difference between our operator and the other one together with its 95\% interval. The runs are denoted by C (operator comparison), T (baseline tuning), and S (separate run on the same images and targets).}
\label{tab:paired}
\centering
\scriptsize
\setlength{\tabcolsep}{3.0pt}
\begin{tabular}{l c c c c c c l l}
\toprule
operator & run & successes & both & ours only & other only & neither & exact $p$ & difference [95\% interval] \\
\midrule
rank-4 Gaussian (ours) & C & 92 & -- & -- & -- & -- & -- & -- \\
\midrule
concept mean, tiled & C & 3 & 3 & 89 & 0 & 58 & $3.2\times10^{-27}$ & $+0.593$ $[+0.515,+0.672]$ \\
GCS direction & C & 8 & 7 & 85 & 1 & 57 & $2.3\times10^{-24}$ & $+0.560$ $[+0.478,+0.642]$ \\
residual steering & T & 1 & 1 & 91 & 0 & 58 & $8.1\times10^{-28}$ & $+0.607$ $[+0.528,+0.685]$ \\
text direction & T & 48 & 39 & 53 & 9 & 49 & $1.1\times10^{-8}$ & $+0.293$ $[+0.202,+0.385]$ \\
VISTA-SAE$^\dagger$ & S & 41 & 39 & 53 & 2 & 56 & $8.6\times10^{-14}$ & $+0.340$ $[+0.260,+0.420]$ \\
AdaIN & C & 27 & 27 & 65 & 0 & 58 & $5.4\times10^{-20}$ & $+0.433$ $[+0.354,+0.513]$ \\
WCT, rank 4 & C & 46 & 45 & 47 & 1 & 57 & $3.5\times10^{-13}$ & $+0.307$ $[+0.231,+0.383]$ \\
WCT, rank 16 & C & 51 & 50 & 42 & 1 & 57 & $1.0\times10^{-11}$ & $+0.273$ $[+0.200,+0.347]$ \\
real donor object & C & 75 & 64 & 28 & 11 & 47 & $9.5\times10^{-3}$ & $+0.113$ $[+0.034,+0.193]$ \\
\bottomrule
\end{tabular}
\end{table}

\section{Norm Matching}
\label{app:norm}
The token written at position $p$ is $s_p x_p$, with $s_p=\lVert h_p\rVert/\lVert x_p\rVert>0$, which has two exact consequences. First, for the tiled mean, the written set is $\{s_p\mu\}$, whose tokens are collinear, and its variation is thus restricted to the direction of $\mu$. The failure of this control (0.020) therefore shows that variance along the mean direction does not, by itself, make the mean readable. Second, for Gaussian draws, the written set remains in $\mathrm{span}\{\mu,v_1,\ldots,v_r\}$, where $v_i$ denote the stored eigenvectors. Rescaling can therefore reweight the draws but cannot create a new direction.

We measure this effect on the 150 discovery images for the rank-4 Gaussian and report medians over the images. The per-patch scales vary considerably, with a coefficient of variation of 0.45. Nevertheless, the cosine between the mean of the written set and $\mu$ is 0.994 before and 0.984 after norm matching, and the share of variance along $\mu$ is 0.19 in both cases. The error of the trace-normalized covariance in the complement of $\mu$, measured w.r.t.\ $\Sigma_4$, is 0.14 before norm matching, which corresponds to the finite-sample error of $m$ draws, and 0.23 after. In Table~\ref{tab:norm}, we repeat the main comparison with one common scale per image, in which case the tiled mean has exactly zero covariance, and without any rescaling. The results show that the normalization does not affect the outcome. In particular, the Gaussian yields 0.593 to 0.600 and the tiled mean 0.013 to 0.020 under all three normalizations, and the outcome of the Gaussian differs from that of the main setting on 11 and 12 of the 150 images, respectively (6:5 and 6:6).

\begin{table}[H]
\caption{Injection under three normalizations of the substituted set, $n=150$.}
\label{tab:norm}
\centering
\small
\begin{tabular}{l c c}
\toprule
normalization & tiled mean & rank-4 Gaussian \\
\midrule
per patch (main setting) & 0.020 & 0.600 \\
one common scale per image & 0.020 & 0.593 \\
none & 0.013 & 0.600 \\
\bottomrule
\end{tabular}
\end{table}

\section{Higher-Order Structure}
\label{app:nongauss}
To vary the higher-order structure while keeping the first two moments fixed, we draw the four latent coordinates of the concept bank from non-Gaussian distributions with zero mean, unit variance, and no correlation. The population mean and covariance are thus those of the rank-4 Gaussian in each case. In particular, we use Rademacher coordinates, which yield 16 distinct vectors (excess kurtosis $-2$), uniform coordinates (excess kurtosis $-1.2$), and Student-$t_5$ coordinates (excess kurtosis $+6$). On the 150 discovery images, these yield injection rates of 0.547, 0.620, and 0.540, respectively, whereas Gaussian coordinates yield 0.600 in the same run. None of the three differs significantly from the Gaussian (17:9, $p=0.17$; 6:9, $p=0.61$; 16:7, $p=0.093$, where the counts denote the images on which the Gaussian succeeds while the other fails, and conversely). Furthermore, all three remain far above each set that lacks the covariance of the target (0.020 to 0.447). The largest difference is 0.060, with a 95\% interval of $[-0.002,+0.122]$.

In the same run, combining the mean of the concept with the pooled covariance truncated to rank four yields 0.447, which is below the rate obtained with the per-class covariance at the same rank (36:13, $p=1.4\times10^{-3}$).

\section{Fidelity of Untouched Content}
\label{app:fidelity}
The retention over all captions mixes two effects, since an edit that fails leaves the caption unchanged and thus obtains a high score. In Table~\ref{tab:fidelity}, we therefore report two additional quantities. The share of \emph{objects kept} is a fact-level measure, which is computed on the 49 images whose unedited caption names an annotated COCO object other than the source and the target, and which is defined as the share of these objects that the edited caption still names. Furthermore, the number of \emph{unsupported objects} is the number of COCO categories per caption that the edited caption names and that are neither annotated in the image, nor the target, nor named in the unedited caption.

The results in Table~\ref{tab:fidelity} lead to two observations. First, substitution affects the untouched content more than addition or recoloring does. In particular, the Gaussian keeps 0.61 of the untouched objects named, a real donor object 0.63, WCT 0.70, and the text direction 0.77, whereas the tiled mean, which almost never inserts, keeps 0.81. The cost of the Gaussian is thus the same as that of a real object. Second, the Gaussian introduces as few unsupported objects as the baselines (0.22 per caption), whereas a real donor object, which carries context from the image it was taken from, introduces 0.39.

\begin{figure}[H]
    \centering
    \includegraphics[width=0.62\linewidth]{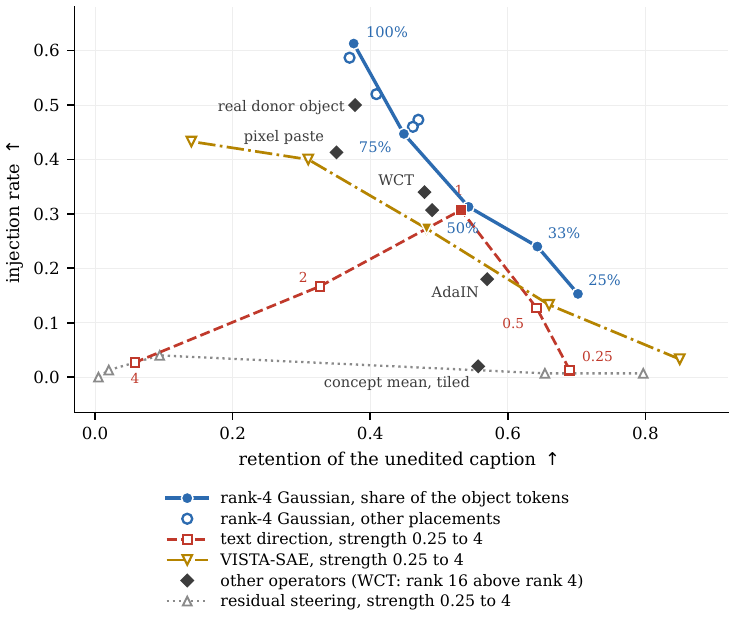}
    \caption{Injection rate against the retention of the unedited caption on the 150 discovery images. For the rank-4 Gaussian, the line connects the settings in which it is written into a share of the object tokens, and the open circles denote the other placements of Section~4.7. For the baselines, the lines connect the swept strengths of Table~\ref{tab:sweep}, and the filled markers denote the selected strengths.}
    \label{fig:fidelity}
\end{figure}

The middle block of Table~\ref{tab:fidelity} traces the injection against the retention for the Gaussian, either by changing the positions at which it is written or by writing it into a share of the object tokens. We observe that the Gaussian outperforms all other operators up to a retention of 0.49, is on par with the best of them at 0.54, and outperforms them at higher retention. In particular, between 0.45 and 0.49 retention, it yields injection rates of 0.45 to 0.47, whereas WCT and VISTA-SAE yield 0.27 to 0.34. At a retention of 0.54, which is reached by writing it into half of the object tokens, it yields 0.313, whereas the text direction yields 0.320 at a retention of 0.535. Furthermore, when it is written into a third and a quarter of the object tokens, it yields 0.240 and 0.153 at retentions of 0.643 and 0.702, respectively. At the comparable retentions of 0.642 and 0.690, the text direction yields 0.127 and 0.013 (Table~\ref{tab:sweep}). The Gaussian thus exceeds the text direction at these two retentions (28:11, $p=9.5\times10^{-3}$, and 22:1, $p=5.7\times10^{-6}$, respectively), whereas the two are on par at 0.54 (24:23 w.r.t.\ the run of Table~\ref{tab:sweep}). Figure~\ref{fig:fidelity} shows the corresponding curves. Finally, we paste the object in pixel space. To this end, the box of a donor object is cropped, resized to the target box, and pasted into the photograph without blending, and no activation is edited. This yields an injection rate of 0.413 at a retention of 0.351, which is below that of the Gaussian in the same run (0.600; 40:12, $p=1.3\times10^{-4}$).

\begin{table}[H]
\caption{Insertion and fidelity on the 150 discovery images. The upper block is obtained from the operator-comparison and baseline-tuning runs, and the middle block from the placement run of Section~4.7 and from a further run with partial coverage. The last row corresponds to pasting the donor crop into the photograph. Dashes denote quantities that were not computed for the respective run.}
\label{tab:fidelity}
\centering
\footnotesize
\setlength{\tabcolsep}{3.5pt}
\begin{tabular}{l c c c c}
\toprule
operator & inject $\uparrow$ & retain $\uparrow$ & \makecell{objects\\kept $\uparrow$} & \makecell{unsupported\\objects $\downarrow$} \\
\midrule
concept mean, tiled & 0.020 & 0.557 & 0.806 & 0.187 \\
AdaIN & 0.180 & 0.570 & 0.820 & 0.173 \\
VISTA-SAE & 0.273 & 0.482 & -- & -- \\
WCT, rank 4 & 0.307 & 0.490 & 0.701 & 0.200 \\
text direction & 0.320 & 0.535 & 0.765 & 0.227 \\
WCT, rank 16 & 0.340 & 0.479 & 0.711 & 0.227 \\
real donor object & 0.500 & 0.378 & 0.633 & 0.393 \\
rank-4 Gaussian & 0.613 & 0.376 & 0.609 & 0.220 \\
\midrule
Gaussian, annotated box & 0.587 & 0.370 & 0.629 & 0.280 \\
Gaussian, centered square, same area & 0.520 & 0.409 & 0.646 & 0.213 \\
Gaussian, random square, same area & 0.473 & 0.470 & 0.629 & 0.253 \\
Gaussian, fixed $8\times8$ square & 0.460 & 0.462 & 0.701 & 0.227 \\
Gaussian, 75\% of the box tokens & 0.447 & 0.449 & 0.687 & 0.153 \\
Gaussian, 50\% of the box tokens & 0.313 & 0.543 & 0.769 & 0.173 \\
Gaussian, 33\% of the box tokens & 0.240 & 0.643 & -- & -- \\
Gaussian, 25\% of the box tokens & 0.153 & 0.702 & -- & -- \\
\midrule
pixel paste of the donor crop & 0.413 & 0.351 & 0.612 & 0.333 \\
\bottomrule
\end{tabular}
\end{table}

Finally, Table~\ref{tab:sweep} reports the injection and the retention of the baselines at each strength that we swept. Note that each strength is evaluated on all 150 discovery images. For the text direction at a strength of 1, this run yields 0.307 at a retention of 0.532, whereas the run of Table~\ref{tab:operators-full} yields 0.320 at 0.535 and a further run yields 0.313 at 0.537. For VISTA-SAE, strengths of 2 and 4 yield higher injection rates at retentions of 0.310 and 0.140, which are below the floor of 0.35 that we use to select the strength of each baseline (Appendix~\ref{app:baselines}).

\begin{table}[H]
\caption{Injection and retention of the insertion baselines at each swept strength, measured on the 150 discovery images. The strength selected for each baseline is marked with $\ast$.}
\label{tab:sweep}
\centering
\small
\begin{tabular}{l c c c c c c}
\toprule
 & & \multicolumn{5}{c}{strength} \\
operator & & 0.25 & 0.5 & 1 & 2 & 4 \\
\midrule
text direction & inject $\uparrow$ & 0.013 & 0.127 & 0.307$^\ast$ & 0.167 & 0.027 \\
 & retain $\uparrow$ & 0.690 & 0.642 & 0.532$^\ast$ & 0.327 & 0.058 \\
\midrule
residual steering & inject $\uparrow$ & 0.007$^\ast$ & 0.007 & 0.040 & 0.013 & 0.000 \\
 & retain $\uparrow$ & 0.797$^\ast$ & 0.654 & 0.094 & 0.020 & 0.005 \\
\midrule
VISTA-SAE & inject $\uparrow$ & 0.033 & 0.133 & 0.273$^\ast$ & 0.400 & 0.433 \\
 & retain $\uparrow$ & 0.850 & 0.660 & 0.482$^\ast$ & 0.310 & 0.140 \\
\bottomrule
\end{tabular}
\end{table}

\textbf{Human rating of the scene.} In addition, two annotators, who were blind to the condition, rated the edited captions that name the target, namely, the 92 captions of the rank-4 Gaussian and the 75 captions of the real donor object in Table~\ref{tab:fidelity}. For each caption, they answered three questions w.r.t.\ the photograph. The first question is whether the caption places the target in the photographed scene, i.e., together with the setting or the objects that are visible, as opposed to describing it alone or in a different scene. The second question is whether the statements about everything other than the target are correct, which is scored with 2 if all of them are correct, with 1 in the case of one error, and with 0 otherwise. The annotators agree on the first question for 94.6\% of the captions ($\kappa=0.61$) and give the same score on the second question for 76.0\%. Averaged over the two annotators, the target is placed in the photographed scene in 0.940 of the captions of the Gaussian and in 0.907 of those of the donor object. Furthermore, the rest of the description receives a mean score of 1.20 and 1.12, respectively, and both annotators give the full score to 0.370 and 0.307 of the captions. The Gaussian is thus on par with a real object in both respects. The third question is whether the count, the attributes and the relation to the scene that the caption gives for the target are plausible for the photograph, where a caption that names the target without describing it is recorded separately. The annotators agree on 94.0\% of the captions ($\kappa=0.86$). The description is rated plausible in 0.826 of the cases for the Gaussian and in 0.587 for the donor object, and implausible in 0.120 and 0.327, respectively, whereas 0.054 and 0.087 of the captions name the target without describing it. Both annotators rate the description as plausible for 0.793 and 0.560 of the captions, respectively.

\end{document}